\documentclass[runningheads]{llncs}
\usepackage{graphicx}
\usepackage{multirow}
\usepackage{float}
\usepackage[para]{threeparttable}
\usepackage{adjustbox}
\usepackage{amsmath}
\usepackage{bm}
\newcommand{\etal}{\textit{et al.}}
\usepackage{arydshln}
\usepackage{amssymb}

\usepackage{color}
\usepackage[misc]{ifsym}

\newcommand\blfootnote[1]{%
  \begingroup
  \renewcommand\thefootnote{}\footnote{#1}%
  \addtocounter{footnote}{-1}%
  \endgroup
}

\begin{document}
\title{Unsupervised Representation Learning Meets Pseudo-Label Supervised Self-Distillation:\\A New Approach to Rare Disease Classification}
%\title{Contribution Title\thanks{Supported by organization x.}}
%
\titlerunning{A Hybrid Representation Learning Approach to Rare Disease Classification}
% If the paper title is too long for the running head, you can set
% an abbreviated paper title here
%
%\author{First Author\inst{1}\orcidID{0000-1111-2222-3333} \and
%Second Author\inst{2,3}\orcidID{1111-2222-3333-4444} \and
%Third Author\inst{3}\orcidID{2222--3333-4444-5555}}
%
\author{Jinghan Sun\inst{1,2} \and
Dong Wei\inst{2} \and
Kai Ma\inst{2} \and
Liansheng Wang\inst{1}\textsuperscript{(\Letter)} \and\\
Yefeng Zheng\inst{2}}
\authorrunning{J. Sun et al.}
% First names are abbreviated in the running head.
% If there are more than two authors, 'et al.' is used.
%
%\institute{Princeton University, Princeton NJ 08544, USA \and
%Springer Heidelberg, Tiergartenstr. 17, 69121 Heidelberg, Germany
%\email{lncs@springer.com}\\
%\url{http://www.springer.com/gp/computer-science/lncs} \and
%ABC Institute, Rupert-Karls-University Heidelberg, Heidelberg, Germany\\
%\email{\{abc,lncs\}@uni-heidelberg.de}}
%
\institute{Xiamen University, Xiamen, China\\
\email{jhsun@stu.xmu.edu.cn}, \email{lswang@xmu.edu.cn} \and
Tencent Jarvis Lab, Shenzhen, China\\
\email{\{donwei,kylekma,yefengzheng\}@tencent.com}}
\maketitle              % typeset the header of the contribution
\begin{abstract}
Rare diseases are characterized by low prevalence and are often chronically debilitating or life-threatening\blfootnote{\textsuperscript{*} J. Sun and D. Wei---Contributed equally; J. Sun contributed to this work during an internship at Tencent.}.
Imaging-based classification of rare diseases is challenging due to the severe shortage in training examples.
Few-shot learning (FSL) methods tackle this challenge by extracting generalizable prior knowledge from a large base dataset of common diseases and normal controls, and transferring the knowledge to rare diseases.
Yet, most existing methods require the base dataset to be labeled and do not make full use of the precious examples of the rare diseases.
To this end, we propose in this work a novel hybrid approach to rare disease classification, featuring two key novelties targeted at the above drawbacks.
First, we adopt the unsupervised representation learning (URL) based on self-supervising contrastive loss,
%present a baseline approach based on unsupervised representation learning (URL) using self-supervising contrastive loss,
whereby to eliminate the overhead in labeling the base dataset.
Second, we integrate the URL with pseudo-label supervised classification for effective self-distillation of the knowledge about the rare diseases, composing a hybrid approach taking advantages of both unsupervised and (pseudo-) supervised learning on the base dataset.
% in the absence of any label of the auxiliary dataset.
Experimental results on classification of rare skin lesions show that our hybrid approach substantially outperforms existing FSL methods (including those using fully supervised base dataset) for rare disease classification via effective integration of the URL and pseudo-label driven self-distillation, thus establishing a new state of the art.
%the baseline approach using the URL alone already outperforms existing FSL methods (including those using fully supervised base dataset) for rare disease classification, and that the hybrid approach further boosts the performance substantially by effective integration of the URL and pseudo-label driven self-distillation.

\end{abstract}
\section{Introduction}
Rare diseases are a significant public health issue and a challenge to healthcare.
On aggregate, the number of people suffering from rare diseases worldwide is estimated over 400 million, and there are about 5000--7000 rare diseases---with 250 new ones appearing each year \cite{stolk2006rare}.
%Research and development of treatments for rare diseases have stagnated because there are too few patients for clinical trials.
Patients with rare diseases face delayed diagnosis: 10\% of patients spent 5--30 years to reach a final diagnosis.
Besides, many rare diseases can be misdiagnosed. % as common diseases
Therefore, image-based accurate and timely diagnosis of rare diseases can be of great clinical value.
In recent years, deep learning (DL) methods have developed into the state of the art (SOTA) for image-based computer-aided diagnosis (CAD) of many diseases \cite{ker2017deep,litjens2017survey,shen2017deep}.
However, due to the limited number of patients for a specific rare disease, collecting sufficient data for well training of generic DL classification models can be practically difficult or even infeasible for rare diseases.
%which means previous deep learning methods for image-based classification of rare diseases usually suffer from scarce data.

To cope with the scarcity of training samples, a machine learning paradigm called few-shot learning (FSL) has been proposed \cite{fei2006one} and achieved remarkable advances in the natural image domain \cite{finn2017model,hsu2018unsupervised,khodadadeh2019unsupervised,snell2017prototypical,vinyals2016matching}.
In FSL, generalizable prior knowledge is learned on a large dataset of base classes, and subsequently utilized to boost learning of previously unseen novel classes given limited samples (the target task).
%FSL aims to recognize a set of classes learn from a limited number of labeled instances per class.
Earlier approaches \cite{finn2017model,hsu2018unsupervised,khodadadeh2019unsupervised,snell2017prototypical,vinyals2016matching} to FSL mostly resorted to the concept of meta-learning
%and adopted the episodic training scheme \cite{vinyals2016matching}.
%{\color{blue}Notably, Li \etal~\cite{li2020difficulty} proposed a difficulty-aware meta-learning method based on the model-agnostic meta-learning (MAML) framework \cite{finn2017model} for rare disease classification.}
%Motivated by model-agnostic meta-learning (MAML), Li \etal \cite{li2020difficulty} develop a difficulty-aware method that meta-training a network with dedicated tasks on labeled data from common diseases to diagnose rare diseases with limited labeled examples.
%However, these approaches often involve
and involved complicated framework design and task construction.
Recently, Tian \etal~\cite{tian2020rethinking} showed that superior FSL performance could be achieved by simply learning a good representation on the base dataset using basic frameworks, followed by fitting a simple classifier to few examples of the novel classes.
Additional performance boosts were achieved through self-distillation \cite{furlanello2018born,hinton2015distilling}.
How to implement a similar self-distilling strategy on an unsupervised base dataset, though, is not obvious.
%In addition, many recent studies \cite{chen2019closer,tian2020rethinking} have shown that training an embedding model followed by a linear classifier outperform state-of-art meta-learning methods with a large margin.
%However, Most of the works require large labeled base classes (common diseases).
On the other hand, we are aware of only few methods \cite{zhu2020alleviating,jiang2019task,li2020difficulty,paul2021discriminative} for FSL of medical image classification, and to the best of our knowledge, all of them relied on heavy labeling of the base dataset, causing a great burden for practical applications.
Lastly, the meta-learning process and the target task are often isolated in most existing FSL approaches,
% (both for natural and medical images)
and the meta-learner has little knowledge about its end task.
For natural images, this setting is consistent with the general purpose of pretraining a classifier that can be quickly adapted for diverse tasks.
For the scenario we consider, however, known types of rare diseases are mostly fixed, and their recognition constitutes a definite task.
%They ignore the available precious knowledge about rare diseases in the test stage and do not leverage the relationship between common diseases and rare diseases.
We hypothesize that, by bridging the base dataset and the definite task,
%and a more targeted learning on the large base dataset
the performance can be boosted for the rare disease classification.
%{\color{blue}We propose that a superior feature extractor can improve the classification accuracy.}
In this regard, we consider this work a specific and practically meaningful application of the general FSL.

In this work, we propose a novel hybrid approach to rare disease classification, which combines unsupervised representation learning (URL) and pseudo-label supervised self-distillation \cite{furlanello2018born,hinton2015distilling}.
Motivated by the recent surge of representation learning in FSL \cite{chen2019closer,tian2020rethinking}, we first build a simple yet effective baseline model based on URL, where a good representation is learned on a large unlabeled base dataset consisting of common diseases and normal controls (CDNC) using contrastive learning \cite{he2020momentum}, and applied to rare disease classification.
%The baseline model is simple yet effective---in fact more effective than many SOTA FSL methods that require full labeling of the base dataset.
So far as we are aware of, this is the first study that explores few-shot medical image classification using an unsupervised base dataset.
%new paradigm of few-shot learning based on unsupervised representation learning (URL) without any labels and human-designed tasks.
Then, we further propose to inject knowledge about the rare diseases into representation learning, to exploit the CDNC data for a more targeted learning of the rare diseases.
Specifically, we use the baseline model as a teacher model to generate pseudo labels for instances of CDNC \textit{belonging to the rare diseases}, to supervise knowledge distillation
%(extracted from the precious examples by the baseline model)
to the student model.
%for injection of the knowledge contained in the precious, limited examples of the rare diseases into the representation learning process.
Our rationale is that CDNC and rare diseases often share common characteristics, %(\textit{e.g.}, color, shape, and/or texture),
thus we can steer the representation learning on the former towards characteristics that can better distinguish the latter via supervision by the pseudo labels.
%Furthermore, the pseudo-label supervised self-distillation integrated with URL can be leverage to mine the potential knowledge of the rare diseases from unlabeled archive data of common diseases since we found that certain features of the data in the common diseases and the rare diseases are overlapped (e.g., color, shape and structure).
In addition, we empirically explore design options for the distillation, and find that a hybrid self-distillation integrating URL and pseudo-label supervised classification yields the best performance.
%, {\color{red}and that the adaptive hard pseudo label beats three alternatives}.
%
%{\color{red}In summary, our contributions are three-fold:
%\begin{itemize}
%  \item We propose a simple yet effective approach to rare disease classification, which is based on URL and eliminates the need for labeling the large archive of CDNC data.
%  \item We further propose to integrate the URL with pseudo-label supervised self-distillation, composing a hybrid approach taking advantages of both methodologies.
%  \item We present a novel design of adaptive pseudo labels for optimal classification performance of the rare diseases.
%\end{itemize}}
%
Therefore, our main contributions are two folds: the novel application of URL to rare disease classification, and hybrid distillation combining contrastive and pseudo-label learning.
Experimental results on the ISIC 2018 skin lesion classification dataset show that the URL-based baseline model already outperforms previous SOTA FSL methods (including those using fully supervised base dataset), and that further boosts in performance are achieved by the proposed hybrid approach.

\begin{figure}[t]
\centering
\includegraphics[width=\textwidth]{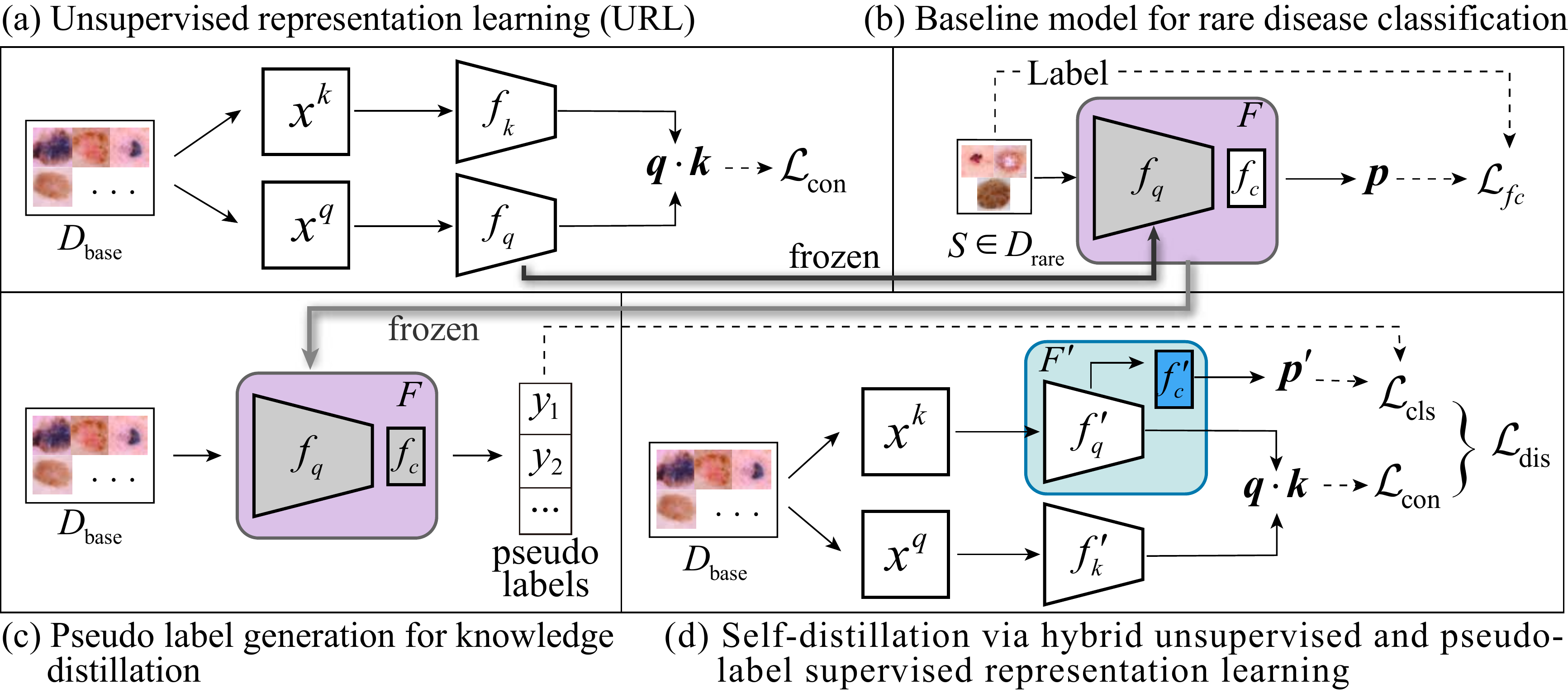}
\caption{Overview of the proposed approach.
Solid line: information flow; dashed line: loss computation.
Note that $\mathcal{L}_{f_c}$ in (b) can be any loss suitable for the classifier $f_c$.}\label{fig:procedure}
\end{figure}

\section{Methods}
\subsubsection{Problem Setting.}\label{sec:methods:state}
Similar to Li \etal~\cite{li2020difficulty}, we formulate the task of rare disease classification as a few-shot learning (FSL) problem.
Specifically, we model a specific task as $\mathcal{T}=\{S, Q\}$ consisting of a support set $S=\{(x, y)\}$ and a query set $Q=\{(x, y)\}$, where $x$ is an image and $y$ is its label.
An $N$-way $K$-shot task includes $N$ rare diseases, each with $K$ instances in $S$, where $K$ is small.
Thus $y\in[1, \ldots, N]$ and $|S|=N\times K$.
The task instance $\mathcal{T}$ is randomly drawn from a distribution $p(\mathcal{T})$, with $S$ and $Q$ randomly sampled from a dataset $D_\mathrm{rare}$ consisting of rare diseases.
Only $S$ is available for training and $Q$ is solely for testing.
In addition, there is a large base dataset $D_\mathrm{base}$ consisting of common diseases and normal controls (CDNC).
The target is optimal classification performance on $Q$ given $S$ and $D_\mathrm{base}$.
In this work, we consider $D_\mathrm{base}$ to be unlabeled for a more generally applicable approach in practice by eliminating the need for annotation.
%{\color{red}However, we still compare the performance with $D_\mathrm{base}$ fully labeled in our experiments for a more comprehensive evaluation.}

\subsubsection{Method Overview.}
An overview of our method is shown in Fig. \ref{fig:procedure}.
Given $D_\mathrm{base}$, we first perform unsupervised representation learning (URL) to train the embedding function $f_q$ (Fig. \ref{fig:procedure}(a)).
Next, a simple classifier $f_c$ is appended to the learned $f_q$ (with frozen parameters) to compose a baseline model $F$ (Fig. \ref{fig:procedure}(b)), where $f_c$ is optimized on $S$.
Then, $F$ is employed to assign each CDNC instance in $D_\mathrm{base}$ a pseudo label of the rare diseases (Fig. \ref{fig:procedure}(c)).
Lastly, a self-distillation via hybrid unsupervised and pseudo-label supervised representation learning is performed on $D_\mathrm{base}$ to produce the final (student) model $F'$ (Fig. \ref{fig:procedure}(d)).
%to make the best classification models, we first ingest the embeddings from the support set $S$ of rare diseases to train a classifier as the teacher model. Then, unsupervised representation learning and self-distillation was used to adapt the student model to acquire ability of feature extraction and learn the relationship between common diseases and rare diseases. Both parts promote each other during training. Lastly, the student model can predict rare diseases classes without any alterations.
% In this section, we first describe in Section \ref{sec:methods:state} the problem setting about the few-shot learning and introduce in Section \ref{sec:method:rep} the proposed method.

%\subsubsection{Unsupervised Representation Learning on Common Diseases and Normal Controls for Rare Disease Classification}
\subsubsection{URL on CDNC for Rare Disease Classification.} \label{sec:method:rep}
%Earlier approaches to few-shot image classification resorted to sophisticated meta-learning algorithms and achieved significant advances \cite{finn2017model,snell2017prototypical,vinyals2016matching}.
%The most recent approaches \cite{finn2017model,snell2017prototypical,vinyals2016matching} for few-shot image classification is training a network in the way of episode training, which is rigid and carefully designed strategy is hard to implement (e.g., number of meta-training tasks, outer learning rate, inner learning rate), thus produce a suboptimal generalization performance on the novel tasks.
%Recently, Tian \etal~\cite{tian2020rethinking} showed that simply learning an embedding representation on the meta-training set, followed by training a linear classier on top of this representation, outperformed state-of-the-art meta-learning based few-shot learning methods, often by large margins.
%, on several benchmark natural image datasets for few-shot classification
Inspired by the recent success of representation learning in FSL \cite{chen2019closer,tian2020rethinking} and based on the recent advances in URL \cite{chen2020simple,he2020momentum}, we propose to perform URL on the big yet unlabeled CDNC dataset for rare disease classification.
Specifically, we adopt instance discrimination with the contrastive loss \cite{he2020momentum,oord2018representation} as our self-supervising task.
% for URL on $D_\mathrm{base}$
We employ MoCo\_v1 \cite{he2020momentum}, where each image $x_i$ in $D_\mathrm{base}$ is augmented twice to obtain $x_i^q$ and $x_i^k$, whose embedded representations are subsequently obtained by $\bm{q}_i=f_q(x_i^q; \theta_q)$ and $\bm{k}_i=f_k(x_i^k; \theta_k)$, where $f_{q}$ and $f_{k}$ are the query and key encoders parameterized by $\theta_q$ and $\theta_k$, respectively.
%Thus we obtain an embedding function $f_q$ with $D_{base}=\{x\}$ via MoCo \cite{he2020momentum}, a contrastive unsupervised learning method. Concretely, each image $x$ is augmented twice into two different samples, and extract its features $f_q(x_q)$, $f_k(x_k)$ respectively.
The contrastive loss $\mathcal{L}_\mathrm{con}$ is defined as \cite{oord2018representation}:
\begin{equation}\label{eq:lcon}
\mathcal{L}_\mathrm{con}(x_i)=-\log\Big[ {\exp \left(\bm{q}_i \cdot \bm{k}_i / \tau\right)} \big/
\big({\exp \left(\bm{q}_i \cdot \bm{k}_i / \tau\right) + {\sum}_{j=1}^{L} \exp \left(\bm{q}_i \cdot \bm{k}_j / \tau\right)}\big) \Big],
\end{equation}
where $L$ is the number of keys stored in the dynamic dictionary implemented as a queue, $\tau$ is a temperature hyperparameter.
Intuitively, this loss is the log loss of an $(L+1)$-way softmax-based classifier trained to discriminate the (augmented) instance $x_i$ from other images stored in the dictionary queue (represented by their embeddings).
Then, $\theta_q$ is updated by back-propagation whereas $\theta_k$ is updated with a momentum $m$: $\theta_k\leftarrow m\theta_k + (1-m)\theta_q$, where $m\in[0, 1)$.
%\textcolor{red}{Please see \cite{he2020momentum} for more details}.
%A contrastive loss $\mathcal{L}_{q}$ function makes the feature distance of the two images from the same original image $f_k(x_{k_+})$ (called positive pair) closer, and at the same time, dissimilar to the other images (called negative pairs).
Another notable difference of our work from the prevailing meta-learning based approaches is that we randomly sample mini-batches of training images from $D_\mathrm{base}$, instead of iteratively constructing episodic training tasks as in most previous works \cite{li2020difficulty,finn2017model,sung2018learning,snell2017prototypical,khodadadeh2019unsupervised,hsu2018unsupervised}.
%Instead of constructing $D_{base}$ in the identical form as the $D_{novel}$ in previous works \cite{finn2017model,snell2017prototypical}, we merge training episodes into a single training set and our $D_{base}$ is unlabeled.
This back-to-basic training scheme has proven effective despite being simpler \cite{tian2020rethinking}, especially for an unsupervised $D_\mathrm{base}$ where category information is missing for episode construction.

After the URL, we freeze $\theta_q$ and append a simple classifier $f_c$ to $f_q$ to form a baseline model $F=f_c(f_q)$ for rare disease classification.
Like \cite{tian2020rethinking}, we use logistic regression for $f_c$, whose parameters $\theta_c$ are optimized on the support set $S$.
%with {\color{red}$f_q$} frozen as a fixed embedding function.
%Given the embedding function {\color{red}$f_q$}, previous studies suggested freezing the extractor and learning a logistic regression classifier with support set $S$.
%As we will show later, $F$ can already achieve a decent baseline performance.

\subsubsection{Self-Distillation of Rare Disease Knowledge.}\label{sec:method:distill}% via Hybrid Unsupervised and Pseudo-Label Supervised Representation Learning
%Previous published studies of supervised learning \cite{tian2020rethinking} has tend to remove the task-specific top layers and take the bottom layers as a feature extractor. In unsupervised representation learning, Researchers usually combine the fixed embedding function with a new linear classifier training from scratch in test stage. Whereas, they totally ignore the knowledge about the rare diseases in training stage.
Despite its decent performance, the baseline model completely ignores the precious knowledge about target rare diseases contained in the support set $S$ during representation learning.
We hypothesize that a better representation learning for classification of the target rare diseases can be achieved by fully exploiting this knowledge while at the same time utilizing the big unlabeled data in $D_\mathrm{base}$.
To do so, we propose to inject target task knowledge extracted from $S$ into the representation learning process via knowledge distillation \cite{hinton2015distilling}, which can transfer knowledge embedded in a teacher model to a student model.
%via a hybrid self-distillation which integrates the contrastive loss and a novel pseudo-label supervised classification loss.
In addition, we adopt the born-again strategy where the teacher and student models have an identical architecture, for its superior performance demonstrated by Furlanello \etal~\cite{furlanello2018born}.
%Furlanello \etal~\cite{furlanello2018born} showed that an identical architecture for the teacher and student models achieved superior performance.

%Note that our purpose is to predict which category this data $x$ belongs to in the test set $D_{novel}$ instead of the training set $D_{base}$.
The key idea behind our knowledge distillation scheme is that, although $D_\mathrm{base}$ and $D_\mathrm{rare}$ comprise disjoint classes, it is common that certain imaging characteristics (\textit{e.g.}, color, shape and/or texture) of the CDNC data are shared by the rare diseases.
Therefore, it is feasible to learn rare-disease-distinctive representations and classifiers by \textit{training the networks to classify CDNC instances as rare diseases of similar characteristics}.
% Next, the frozen $f_q$ and classifier are chained as teacher model.
% Two kinds of knowledge are distilled at the same time: embedding function and knowledge about the target classes.
Mathematically, we use the baseline model $F$ as the teacher model {to predict the probabilities of each image $x$ in $D_\mathrm{base}$ belonging to the rare diseases in $D_\mathrm{rare}$}: $\bm{p}=F(x)=[p_1, \ldots, p_N]^T$, where ${\sum}^N_{n=1}{p_n}=1$.
Next, we define the pseudo label $\bm{y}=[y_1, \ldots, y_N]^T$ based on $\bm{p}$ with two alternative strategies: (i) hard labeling where $y_n=1$ if $n=\operatorname{argmax}_n p_n$ and 0 otherwise, and (2) soft labeling where $y_n=p_n$.
In effect, the first strategy indicates the rare disease that $x$ resembles most, whereas the second reflects the extents of resemblance between $x$ and all the rare diseases in $D_\mathrm{rare}$.
In addition, we propose a hybrid distilling loss integrating pseudo-label supervised classification and contrastive instance discrimination.
As we will show, the hybrid distillation scheme is important for preventing overfitting to noise and bias in the small support set.
Then, adopting the born-again \cite{furlanello2018born} strategy, a randomly initialized student model $F'=f'_c(f'_q)$ parameterized by $\theta'_c$  and $\theta'_q$ is trained to minimize the hybrid loss $\mathcal{L}_\mathrm{dis}$:
%summing the unsupervised constrastive loss $L_\mathrm{con}$ and pseudo-label supervised classification loss $L_\mathrm{cls}$:
\begin{equation}\label{eq:distil}
    \mathcal{L}_\mathrm{dis} = \mathcal{L}_\mathrm{con}(x; \theta'_q, \theta'_k) + \mathcal{L}_\mathrm{cls}\big(\bm{y}, F'(x; \theta'_q, \theta'_c)\big),
    %(\theta'_q, \theta'_k, \theta'_c) = \operatorname{argmin}_{(\theta'_q, \theta'_k, \theta'_c)}\Big(\mathcal{L}_\mathrm{con}(x; \theta'_q, \theta'_k) + L_\mathrm{cls}\big(\bm{y}, F'(x; \theta'_q, \theta'_c)\big)\Big),
\end{equation}
where $\theta'_k$ are parameters of the key encoder $f'_k$ (for computation of $\mathcal{L}_\mathrm{con}$) and updated along with $\theta'_q$, and $\mathcal{L}_\mathrm{cls}$ is the pseudo-label supervised classification loss.
For $\mathcal{L}_\mathrm{cls}$, the cross-entropy and Kullback-Leibler divergence losses are used for the hard and soft labels, respectively.
%the cross-entropy loss is used for hard labels, whereas the Kullback-Leibler divergence loss is used for soft labels.
% \begin{equation}\label{eq:ce_loss}
%     \mathcal{L}_{cls} =
%     \mathcal{L}_\mathrm{CE}(y^h, f_s(x)) =
%     \mathcal{L}_\mathrm{CE}(\bm{y^h},\bm{p}) =
%     -{\sum}^N_{n=1}{y^h_n}\log{p_n},
% \end{equation}
% \begin{equation}\label{eq:kl_loss}
%     \mathcal{L}_{cls} =
%     \mathcal{L}_\mathrm{KL}(\bm{y^t}, f_s(x)) =
%     \mathcal{L}_\mathrm{KL}(\bm{y^t}||\bm{p}) =
%     {\sum}_{n}y^t_n\log (y^t_n/p_n).
% \end{equation}
Lastly, to allow for an end-to-end training, a fully connected layer (followed by softmax) is used for $f'_c$.
%, in contrast to the logistic regression for $f_c$.

%\subsection{Using Student Model for Rare Disease Classification}
After distillation, the student model $F'=f_c'(f_q')$ can be directly used for rare disease classification.
One might argue for an alternative way of usage: discarding $f'_c$ but appending a logistic regression classifier fit to the support set to $f'_q$, just like in the baseline model.
%we consider two alternative ways to apply the student model to rare disease classification: directly using $f'_c$ for prediction, %(2) finetuing $f'_c$ on $S$ for prediction,
%or discarding $f'_c$ but appending a logistic regression classifier to $f'_q$ with with its parameters frozen (just like in the baseline model).
%(using $f'_q$ as a fixed embedding function).
%directly predict the class of query set in $D_{novel}$ without finetuning during the meta-testing stage.
However, as confirmed by our comparative experiment (supplement Table \ref{tab:classifier}), direct use of $F'$ performs much better.
This is because, the knowledge about the rare diseases is distilled into not only $f'_q$ but also $f'_c$, thus discarding the latter results in performance degradation.
Lastly, through preliminary experiments we find that distilling more than once does not bring further improvement.
Therefore, we perform the self-distillation only once.

\subsubsection{Adaptive Pseudo Labels.}
In practice, the pseudo labels defined by $F$ may not be entirely trustworthy, given the tiny size and potential noise and bias of the support set.
%due to the sampling process $\mathcal{T}\sim p(\mathcal{T})$.
This may adversely affect performance of the student model.
%However, only using the $K$ support samples in $S$ to fit a classifier may output some noisy labels because of noisy samples or biased distributions, which may adversely affect the performance.
To alleviate the adverse effect, we further propose adaptive pseudo labels based on the self-adaptive training \cite{huang2020self} strategy.
Concretely, given the prediction $\bm{p}'$ by the student model and pseudo labels $\bm{y}$ defined above, we combine them as our new training target:
$%\begin{equation}
    \bm{y}^\mathrm{adpt} = (1-\alpha)\times \bm{y} + \alpha\times \bm{p}',
$ %\end{equation}
where $\alpha$ is a confidence parameter controlling how much we trust the teacher's knowledge.
$\bm{y}^\mathrm{adpt}$ is termed adaptive hard/soft labels depending on $\bm{y}$ being hard or soft pseudo labels.
Many previous works used a constant $\alpha$ \cite{huang2020self,zhang2019your}.
In the first few epochs, however, the student model lacks reliability---it gradually develops a better ability of rare disease classification as the training goes on.
Therefore, we adopt a linear growth rate \cite{kim2020self} for $\alpha$ at the $t$\textsuperscript{th} epoch:
$%\begin{equation}\label{eq:lineara}
    \alpha_t = \alpha_T \times (t/T),
$ %\end{equation}
where $\alpha_T$ is the last-epoch value and set to 0.7 as in \cite{kim2020self}, and $T$ is the total number of epochs.
As suggested by the comparative experiments (supplement Table \ref{tab:pseudo_label}), the adaptive hard labels work the best, thus are used in our full model for comparison with other methods.

\section{Experiments}
%In this section, we first describe the datasets and experimental details in Section \ref{sec:exp:setup}. Then, in section \ref{sec:exp:compare}, a comparison with state-of-the-art methods with two different testing scenarios is conducted. Finally, in section \ref{sec:exp:ablation} we present an ablation study of our approach, investigating the effect of network architecture, classifier, and pseudo labeling strategies. To futher confirm the improvements of unsupervised contrastive loss $\mathcal{L}_\mathrm{con}$ and pseudo-label supervised classification loss $\mathcal{L}_\mathrm{cls}$, we evaluate our method based on different combinations of loss functions.

%\subsubsection{Experimental Settings}\label{sec:exp:setup}
\subsubsection{Dataset and Evaluation Protocol.}
The ISIC 2018 skin lesion classification dataset \cite{codella2019skin,tschandl2018ham10000}\footnote{https://challenge2018.isic-archive.com/task3/} includes 10,015 dermoscopic images from seven disease categories: melanocytic nevus (6,705), benign keratosis (1,099),  melanoma (1,113), basal cell carcinoma (514), actinic keratosis (327), dermatofibroma (115), and vascular lesion (142).
From that, we simulate the task of rare disease classification as below.
Following Li \etal~\cite{li2020difficulty}, we use the four classes with the most cases as the CDNC dataset $D_\mathrm{base}$, and the other three as the rare disease dataset $D_\mathrm{rare}$.
%We split classes into two disjoint sets: $D_{base}$ (four classes with the largest number of samples as common diseases), $D_{novel}$ (the left three classes as rare diseases).
$K$ images are sampled for each class in $D_\mathrm{rare}$ to compose the support set $S$ for a 3-way $K$-shot task.
Compared to the binary classification tasks \cite{li2020difficulty}, the multi-way evaluation protocol more genuinely reflects the practical clinical need where more than two rare diseases are present, albeit more challenging.
As to $K$, we experiment with 1, 3, and 5 shots in this work.
All remaining images in $D_\mathrm{rare}$ compose the query set $Q$ for performance evaluation.
Again, this task construction more genuinely reflects the intended scenario of rare disease classification---where only few examples are available for training a classifier to be applied to all future test cases---than the repeated construction of small $Q$'s \cite{li2020difficulty}.
%{\color{red}commonly adopted in the natural image domain}.
We sample three random tasks $\mathcal{T}\sim p(\mathcal{T})$ and report the mean and standard deviation of these tasks.
%For the first testing scenario (S1), we randomly sample 1000 episode tasks in $D_{novel}$ to evaluate our method on the most common setting in the few-shot classification. Based on the situation which approached reality, only $N\times K$ labels are available in $D_{novel}$. We use these labeled samples to train a network in our proposed method or finetuned in another method, and finally test on the rest of samples.
Besides accuracy, we additionally employ the F1 score as the evaluation metric considering the data imbalance between the rare disease classes.

\subsubsection{Implementation.}
The PyTorch \cite{steiner2019pytorch} framework (1.4.0) is used for experiments.
We use the ResNet-12 \cite{lee2019meta,tian2020rethinking,ravichandran2019few} architecture as backbone network for $f_q$ and $f_k$, for its superior performance in the comparative experiments (supplement Table \ref{tab:backbone}).
We train the networks for 200 epochs with a mini-batch size of 16 images on 4 Tesla V100 GPUs.
We adopt the stochastic gradient descent optimizer with a momentum of 0.9 and a weight decay of 0.0001.
The learning rate is initialized to 0.03 and decays at 120 and 160 epochs by multiplying by 0.1.  The feature dimension of the encoded representation is 128, and the number of negatives in the memory bank \cite{he2020momentum} is 1280.
The temperature $\tau$ in Equation (\ref{eq:lcon}) is set to 0.07 as in \cite{he2020momentum}.
%, and $\alpha_T$ in Equation (\ref{eq:lineara}) is set to 0.7 following \cite{kim2020self}.
%For pseudo-label supervised representation learning, we set the $\alpha_T$ in Eqn.(\ref{eq:lineara}) to 0.7 following \cite{kim2020self}.
All images are resized to 224$\times$224 pixels.
Online data augmentation including random cropping, flipping, color jittering, and blurring \cite{chen2020simple} is performed.
The source code is available at: https://github.com/SunjHan/Hybrid-Representation-Learning-Approach-for-Rare-Disease-Classification.

\subsubsection{Comparison to SOTA Methods.}\label{sec:exp:compare}
%The results of our method and others on the task of 3-way 1/3/5-shot image classification on skin lesion dataset are shown in Table \ref{tab_isic}.
According to the labeling status of the base dataset
%(supervised versus unsupervised)
and genre of the methodologies,
%(meta-learning versus representation learning),
all the compared methods are grouped into four quadrants:
(i) supervised meta-learning (SML) including MAML \cite{finn2017model}, Relation Networks \cite{sung2018learning}, Prototypical Networks \cite{snell2017prototypical}, and DAML \cite{li2020difficulty},
(ii) unsupervised meta-learning (UML) including UMTRA \cite{khodadadeh2019unsupervised} and CACTUs \cite{hsu2018unsupervised},
(iii) supervised representation learning (SRL; with or without self-distillation) \cite{tian2020rethinking}, and
(iv) URL including SimCLR~\cite{chen2020simple}, MoCo\_v2~\cite{chen2020improved}, MoCo\_v1~\cite{he2020momentum} (composing the baseline model in our work), and our proposed method.
%For both of the testing scenarios, We compare four different kinds of few-shot learning methods. i) Supervised meta-learning based approaches : MAML \cite{finn2017model}, Relation Networks \cite{sung2018learning}, and Prototypical Networks \cite{snell2017prototypical} ii) unsupervised meta-learning based approaches: UMTRA \cite{khodadadeh2018unsupervised} and CACTUs \cite{hsu2018unsupervised}. iii) Supervised representation learning (SRL) and its with self-distillation (RFS) \cite{tian2020rethinking}. iv) Unsupervised representation learning \cite{chen2020simple,he2020momentum,chen2020improved}.
These methods cover a wide range of the latest advances in FSL for image classification.
For reference purpose, we also show the results of training a classifier from scratch solely on $S$.
Besides the ResNet-12 backbone, we additionally show the results using the 4 conv blocks \cite{vinyals2016matching} as the backbone network considering its prevalent usage in the FSL literature \cite{finn2017model,sung2018learning,snell2017prototypical,khodadadeh2019unsupervised,hsu2018unsupervised}.
Note that for all compared methods, we optimize their performance via empirical parameter tuning.

\begin{table}[t]
\caption{Evaluation results and comparison with SOTA FSL methods with ResNet-12 as backbone network.
Data format: mean (standard deviation).}\label{tab_isic}% of three random tasks
\begin{adjustbox}{width=\columnwidth}
\begin{threeparttable}
\begin{tabular}{l|c|c|c|c|c|c}
\hline
\multirow{2}{*}{Method} & \multicolumn{2}{c}{($N$, $K$) = (3, 1)} &  \multicolumn{2}{|c}{($N$, $K$) = (3, 3)} & \multicolumn{2}{|c}{($N$, $K$) = (3, 5)}  \\
\cline{2-7}
& Accuracy (\%)  & F1 score (\%)  & Accuracy (\%) & F1 score (\%)  & Accuracy (\%)  & F1 score (\%) \\
\hline
Training from scratch  & 37.74 (1.07) & 29.90 (3.65)  & 39.76 (0.88) & 35.60 (1.72)  & 45.36 (3.76) & 38.41 (4.06)  \\
\hdashline
    $\triangleright$ SML
\hfill MAML \cite{finn2017model}   & 47.49 (5.38) & 42.33 (6.16)  & 55.55 (3.12) & 49.19 (4.20)  & 58.94 (2.59) & 53.51 (2.46) \\
\hfill RelationNet \cite{sung2018learning}   & 46.10 (4.80) & 39.98 (6.73)  & 47.29 (2.77) & 43.37 (3.65)  & 55.71 (3.30) & 49.34 (3.57)  \\
\hfill ProtoNets \cite{snell2017prototypical}   & 35.18 (3.12) & 30.81 (3.09)  & 38.59 (1.91) & 33.11 (2.08)  & 42.45 (2.45) & 34.92 (3.70)  \\
\hfill DAML \cite{li2020difficulty}   & 50.05 (5.18) & 41.65 (3.98)  & 55.57 (3.55) & 49.01 (6.62)  & 59.44 (3.17) & 54.66 (2.43)  \\
\hdashline
    $\triangleright$ UML
\hfill UMTRA \cite{khodadadeh2019unsupervised}  & 45.88 (3.63) & 41.44 (4.37)  & 51.29 (3.54) & 45.91 (3.96)  & 57.33 (1.76) & 53.06 (0.89)  \\
\hfill CACTUs-MAML \cite{hsu2018unsupervised}  & 42.98 (2.91) & 35.38 (3.08)  & 44.44 (3.35) & 39.94 (3.65)  & 48.11 (4.20) & 44.32 (3.65)  \\
\hfill CACTUs-ProtoNets \cite{hsu2018unsupervised}   & 42.67 (2.43) & 39.24 (2.72)  & 45.00 (3.26) & 39.69 (2.66)  & 47.95 (3.52) & 44.08 (2.63)  \\
\hdashline
    $\triangleright$ SRL
\hfill SRL-simple \cite{tian2020rethinking}   & 54.45 (5.82) & 51.02 (6.93)  & 61.31 (6.31) & 57.65 (3.46)  & 70.53 (2.17) & 65.58 (3.72)  \\
\hfill SRL-distil \cite{tian2020rethinking}    & 55.43 (7.36) & 51.18 (5.50)  & 64.92 (6.00) & 59.88 (4.87)  & 72.78 (1.67) & 65.89 (2.54)  \\
\hdashline
    $\triangleright$ URL
\hfill SimCLR \cite{chen2020simple}         & 52.43 (5.01) & 44.70 (8.24)  & 63.82 (3.70) & 57.55 (3.67)  & 70.18 (1.76) & 63.73 (1.78)  \\
\hfill MoCo\_v2 \cite{chen2020improved}     & 59.95 (4.73) & 55.98 (3.81)  & 70.84 (2.91) & 64.77 (3.69)  & 75.80 (1.85) & 70.69 (2.13)  \\
\hfill MoCo\_v1 (baseline) \cite{he2020momentum}      & 61.90 (2.92) & 56.30 (1.48)  & 74.92 (2.96) & 69.50 (5.72)  & 79.01 (2.00) & 74.47 (3.03)   \\
\hfill \textbf{Hybrid distil (ours)}  & \textbf{64.15} (2.86) & \textbf{61.01} (1.30) & \textbf{75.82} (2.47) & \textbf{73.34} (2.30) & \textbf{81.16} (2.60) & \textbf{77.35} (4.21) \\
\hline
\end{tabular}
% \begin{tablenotes}
% \item[a] Method grouping: SML: supervised meta-learning, UML: unsupervised meta-learning, SRL: supervised representation learning, and URL: unsupervised representation learning.
% \end{tablenotes}
% \end{threeparttable}
% \begin{tablenotes}\footnotesize
%     \item[a] Despite our efforts, we are unable to tune ProtoNets to yield comparable results with other methods.
% \end{tablenotes}
\end{threeparttable}
\end{adjustbox}
\end{table}

The results are shown in Table \ref{tab_isic} (ResNet-12) and supplement Table \ref{tab_isic_4conv} (4 conv blocks), on which we make the following observations.
First, the representation learning based methods generally achieve better performance than the meta-learning based irrespective of the labeling status of $D_\mathrm{base}$, which is consistent with the findings in the natural image domain \cite{tian2020rethinking}.
%First, SRL is comparable with meta-learning methods indicating that a superior feature extractor is significant for few-shot image classification.
Second, the URL based methods surprisingly outperform the SRL based in most circumstances.
%Surprisingly, we observe that unsupervised approaches perform better than supervised ones.
A possible explanation may be that the few classes in $D_\mathrm{base}$ present limited variations and make the representations overfit to their differentiation, whereas the task of instance discrimination in URL forces the networks to learn more diverse representations that are more generalizable on novel classes.
%A possible explanation for this might be that the inter-class variation in natural images is larger than medical images. So the supervised representation learning methods fail to learn high-quality feature representations from images, whereas unsupervised methods have better generalization ability to capture expressive feature representations.
Especially, the baseline model presented in this work (URL with MoCo\_v1 \cite{he2020momentum}) wins over the SRL plus self-distillation \cite{tian2020rethinking} by large margins.
% in all settings and for both evaluation metrics.
Third, our proposed hybrid approach brings further improvements upon the baseline model in both accuracy ($\sim$1--2\% with ResNet-12) and F1 score ($\sim$3--5\% with ResNet-12).
Notably, it achieves an accuracy of 81.16\% in the 5-shot setting without any label of the base dataset.
These results strongly support our hypothesis that, by extracting the knowledge about the rare diseases from the small support set and injecting it into the representation learning process via pseudo-label supervision, we can fully exploit the large CDNC dataset to learn representations and classifiers that can better distinguish the rare diseases.
To further investigate whether the improvements sustain higher $K$ values, we conduct extra experiments (with ResNet-12) for the baseline and hybrid models with $K=10$ and $20$.
The results confirm that our proposed hybrid approach still wins over the baseline model by $\sim$1\% absolute differences in both metrics and settings, yielding the accuracies and F1 scores of 83.65\% and 79.98\% when $K=10$, and of 86.91\% and 83.40\% when $K=20$.
Lastly, the results using ResNet-12 as backbone are generally better than those using the 4 conv blocks, as expected.

\subsubsection{Ablation Study.}\label{sec:exp:ablation}
%\textit{Effect of Hybrid Distillation.}
We probe the effect of the proposed hybrid distillation by comparing performance of distilling with only $\mathcal{L}_\mathrm{cls}$ and the full model.
In addition, we experiment with a variant of $\mathcal{L}_\mathrm{dis}$ where $\mathcal{L}_\mathrm{cls}$ is replaced by an L1 loss $\mathcal{L}_\mathrm{reg}$ to directly regress the output of $f_q$, to evaluate the effect of injecting knowledge about the rare diseases.
Results (Table \ref{tab:hybrid_distil}) show that distilling with $\mathcal{L}_\mathrm{cls}$ alone brings moderate improvement upon the baseline model, suggesting the efficacy of injecting rare disease knowledge into representation learning.
Yet, distilling with the proposed hybrid loss achieves further appreciable improvement.
We conjecture this is because $\mathcal{L}_\mathrm{con}$ helps avoid overfitting to the support set of the rare diseases, which may be affected by noise and bias due to its small size.
On the other hand, distilling with $\mathcal{L}_\mathrm{reg}$ (plus $\mathcal{L}_\mathrm{con}$) gives performance similar to the baseline model, implying that it is the distilled knowledge about the rare diseases that matters, rather than the distillation procedure.
These results confirm the efficacy of the hybrid distillation.

\begin{table}
    \caption{Ablation study on the hybrid distillation (with ResNet-12 backbone and adaptive hard labels).
    Data format: mean (standard deviation).}\label{tab:hybrid_distil}
    \begin{adjustbox}{width=\columnwidth}
        \begin{tabular}{l|c|c|c|c|c|c}
        \hline
        \multirow{2}{*}{$\mathcal{L}_\mathrm{dis}$} &  \multicolumn{2}{c}{($N$, $K$) = (3, 1)} & \multicolumn{2}{|c}{($N$, $K$) = (3, 3)} & \multicolumn{2}{|c}{($N$, $K$) = (3, 5)} \\
        \cline{2-7}
        & Accuracy (\%)  & F1 score (\%)  & Accuracy (\%) & F1 score (\%)  & Accuracy (\%)  & F1 score (\%) \\
        \hline
        N.A.  & 61.90 (2.92) & 56.30 (1.48)  & 74.92 (2.96) & 69.50 (5.72)  & 79.01 (2.00) & 74.47 (3.03)   \\
        $\mathcal{L}_\mathrm{cls}$  & 63.70 (3.39)& 57.31 (7.73)& 74.92 (2.10)& 70.28 (3.97)& 80.24 (1.61)& 77.29 (2.91)\\
        $\mathcal{L}_\mathrm{con}$ + $\mathcal{L}_\mathrm{cls}$  & \textbf{64.15} (2.86) & \textbf{61.01} (1.30) & \textbf{75.82} (2.47) & \textbf{73.34} (2.30) & \textbf{81.16} (2.60) & \textbf{77.35} (4.21) \\
        $\mathcal{L}_\mathrm{con}$ + $\mathcal{L}_\mathrm{reg}$  & 62.20 (5.18)& 56.19 (4.28)& 74.43 (2.88)& 69.74 (4.13)& 79.14 (2.09)& 74.41 (2.65)\\

        \hline
        \end{tabular}
        \end{adjustbox}
\end{table}

\section{Conclusion}
In this work, we proposed a novel approach to rare disease classification in two steps.
First, we built a baseline model on unsupervised representation learning for a simple and label-free (w.r.t. the base dataset of common diseases and normal controls) framework, which achieved superior performance to existing FSL methods on skin lesion classification.
%combining the recent advances in representation learning for FSL and contrastive loss for unsupervised learning,
Second, we further proposed to utilize the baseline model as the teacher model for a hybrid self-distillation integrating unsupervised contrastive learning and pseudo-label supervised classification.
Experimental results suggested that the hybrid approach could effectively inject knowledge about the rare diseases into the representation learning process through the pseudo-labels, and meanwhile resist overfitting to noise and bias in the small support set thanks to the contrastive learning, and that it had set a new state of the art for rare disease classification.
%Additionally, we noticed an interesting phenomenon that the baseline model which did not use any label of the base dataset outperformed all competing methods that did.
%We plan to investigate more about the underlying reasons.
\subsubsection{Acknowledgments.} This work was supported by the Fundamental Research Funds for the Central Universities (Grant No. 20720190012), Key-Area Research and Development Program of Guangdong Province, China (No. 2018B010111001), and Scientific and Technical Innovation 2030 - ``New Generation Artificial Intelligence'' Project (No. 2020AAA0104100).
%
% ---- Bibliography ----
%
% BibTeX users should specify bibliography style 'splncs04'.
% References will then be sorted and formatted in the correct style.
%
\bibliographystyle{splncs04}
\bibliography{samplepaper}

\begin{thebibliography}{10}
\providecommand{\url}[1]{\texttt{#1}}
\providecommand{\urlprefix}{URL }
\providecommand{\doi}[1]{https://doi.org/#1}

\bibitem{chen2020simple}
Chen, T., Kornblith, S., Norouzi, M., Hinton, G.: A simple framework for
  contrastive learning of visual representations. In: International Conference
  on Machine Learning. pp. 1597--1607. PMLR (2020)

\bibitem{chen2019closer}
Chen, W.Y., Liu, Y.C., Kira, Z., Wang, Y.C.F., Huang, J.B.: A closer look at
  few-shot classification. In: International Conference on Learning
  Representations (2019)

\bibitem{chen2020improved}
Chen, X., Fan, H., Girshick, R., He, K.: Improved baselines with momentum
  contrastive learning. arXiv preprint arXiv:2003.04297  (2020)

\bibitem{codella2019skin}
Codella, N., Rotemberg, V., Tschandl, P., Celebi, M.E., Dusza, S., Gutman, D.,
  Helba, B., Kalloo, A., Liopyris, K., Marchetti, M., et~al.: {Skin Lesion
  Analysis Toward Melanoma Detection 2018: A Challenge hosted by the
  International Skin Imaging Collaboration (ISIC)}. arXiv preprint
  arXiv:1902.03368  (2019)

\bibitem{finn2017model}
Finn, C., Abbeel, P., Levine, S.: Model-agnostic meta-learning for fast
  adaptation of deep networks. In: International Conference on Machine
  Learning. pp. 1126--1135. PMLR (2017)

\bibitem{furlanello2018born}
Furlanello, T., Lipton, Z., Tschannen, M., Itti, L., Anandkumar, A.: Born again
  neural networks. In: International Conference on Machine Learning. pp.
  1607--1616. PMLR (2018)

\bibitem{he2020momentum}
He, K., Fan, H., Wu, Y., Xie, S., Girshick, R.: Momentum contrast for
  unsupervised visual representation learning. In: Proceedings of the IEEE/CVF
  Conference on Computer Vision and Pattern Recognition. pp. 9729--9738 (2020)

\bibitem{hinton2015distilling}
Hinton, G., Vinyals, O., Dean, J.: Distilling the knowledge in a neural
  network. In: NIPS Deep Learning and Representation Learning Workshop (2015)

\bibitem{hsu2018unsupervised}
Hsu, K., Levine, S., Finn, C.: Unsupervised learning via meta-learning. In:
  International Conference on Learning Representations (2018)

\bibitem{Huang_2017_CVPR}
Huang, G., Liu, Z., van~der Maaten, L., Weinberger, K.Q.: Densely connected
  convolutional networks. In: Proceedings of the IEEE Conference on Computer
  Vision and Pattern Recognition (2017)

\bibitem{huang2020self}
Huang, L., Zhang, C., Zhang, H.: Self-adaptive training: Beyond empirical risk
  minimization. arXiv preprint arXiv:2002.10319  (2020)

\bibitem{jiang2019task}
Jiang, X., Ding, L., Havaei, M., Jesson, A., Matwin, S.: Task adaptive metric
  space for medium-shot medical image classification. In: International
  Conference on Medical Image Computing and Computer Assisted Intervention. pp.
  147--155. Springer (2019)

\bibitem{ker2017deep}
{Ker}, J., {Wang}, L., {Rao}, J., {Lim}, T.: Deep learning applications in
  medical image analysis. IEEE Access  \textbf{6},  9375--9389 (2018)

\bibitem{khodadadeh2019unsupervised}
Khodadadeh, S., Boloni, L., Shah, M.: Unsupervised meta-learning for few-shot
  image classification. In: Advances in Neural Information Processing Systems.
  vol.~32. Curran Associates, Inc. (2019)

\bibitem{kim2020self}
Kim, K., Ji, B., Yoon, D., Hwang, S.: Self-knowledge distillation: A simple way
  for better generalization. arXiv preprint arXiv:2006.12000  (2020)

\bibitem{lee2019meta}
Lee, K., Maji, S., Ravichandran, A., Soatto, S.: Meta-learning with
  differentiable convex optimization. In: Proceedings of the IEEE/CVF
  Conference on Computer Vision and Pattern Recognition. pp. 10657--10665
  (2019)

\bibitem{fei2006one}
Li, F.F., Fergus, R., Perona, P.: One-shot learning of object categories. IEEE
  Transactions on Pattern Analysis and Machine Intelligence  \textbf{28}(4),
  594--611 (2006)

\bibitem{li2020difficulty}
Li, X., Yu, L., Jin, Y., Fu, C.W., Xing, L., Heng, P.A.: Difficulty-aware
  meta-learning for rare disease diagnosis. In: International Conference on
  Medical Image Computing and Computer Assisted Intervention. pp. 357--366.
  Springer (2020)

\bibitem{litjens2017survey}
Litjens, G., Kooi, T., Bejnordi, B.E., Setio, A.A.A., Ciompi, F., Ghafoorian,
  M., Van Der~Laak, J.A., Van~Ginneken, B., S{\'a}nchez, C.I.: A survey on deep
  learning in medical image analysis. Medical Image Analysis  \textbf{42},
  60--88 (2017)

\bibitem{oord2018representation}
Oord, A.v.d., Li, Y., Vinyals, O.: Representation learning with contrastive
  predictive coding. arXiv preprint arXiv:1807.03748  (2018)

\bibitem{paul2021discriminative}
Paul, A., Tang, Y.X., Shen, T.C., Summers, R.M.: Discriminative ensemble
  learning for few-shot chest {X}-ray diagnosis. Medical Image Analysis
  \textbf{68},  101911 (2021)

\bibitem{ravichandran2019few}
Ravichandran, A., Bhotika, R., Soatto, S.: Few-shot learning with embedded
  class models and shot-free meta training. In: Proceedings of the IEEE/CVF
  International Conference on Computer Vision. pp. 331--339 (2019)

\bibitem{Sandler_2018_CVPR}
Sandler, M., Howard, A., Zhu, M., Zhmoginov, A., Chen, L.C.: {MobileNetV2}:
  Inverted residuals and linear bottlenecks. In: Proceedings of the IEEE
  Conference on Computer Vision and Pattern Recognition (2018)

\bibitem{shen2017deep}
Shen, D., Wu, G., Suk, H.I.: Deep learning in medical image analysis. Annual
  Review of Biomedical Engineering  \textbf{19}(1),  221--248 (2017)

\bibitem{snell2017prototypical}
Snell, J., Swersky, K., Zemel, R.: Prototypical networks for few-shot learning.
  In: Advances in Neural Information Processing Systems. pp. 4080--4090 (2017)

\bibitem{steiner2019pytorch}
Steiner, B., DeVito, Z., Chintala, S., Gross, S., Paszke, A., Massa, F., Lerer,
  A., Chanan, G., Lin, Z., Yang, E., et~al.: {PyTorch}: An imperative style,
  high-performance deep learning library. Advances in Neural Information
  Processing Systems  \textbf{32},  8026--8037 (2019)

\bibitem{stolk2006rare}
Stolk, P., Willemen, M.J., Leufkens, H.G.: Rare essentials: Drugs for rare
  diseases as essential medicines. Bulletin of the World Health Organization
  \textbf{84},  745--751 (2006)

\bibitem{sung2018learning}
Sung, F., Yang, Y., Zhang, L., Xiang, T., Torr, P.H., Hospedales, T.M.:
  Learning to compare: Relation network for few-shot learning. In: Proceedings
  of the IEEE Conference on Computer Vision and Pattern Recognition (June 2018)

\bibitem{tian2020rethinking}
Tian, Y., Wang, Y., Krishnan, D., Tenenbaum, J.B., Isola, P.: Rethinking
  few-shot image classification: A good embedding is all you need? In:
  Proceedings of the European Conference on Computer Vision (2020)

\bibitem{tschandl2018ham10000}
Tschandl, P., Rosendahl, C., Kittler, H.: The {HAM}10000 dataset, a large
  collection of multi-source dermatoscopic images of common pigmented skin
  lesions. Scientific Data  \textbf{5}(1), ~1--9 (2018)

\bibitem{vinyals2016matching}
Vinyals, O., Blundell, C., Lillicrap, T., Kavukcuoglu, K., Wierstra, D.:
  Matching networks for one shot learning. In: Advances in Neural Information
  Processing Systems. pp. 3637--3645 (2016)

\bibitem{zhang2019your}
Zhang, L., Song, J., Gao, A., Chen, J., Bao, C., Ma, K.: Be your own teacher:
  Improve the performance of convolutional neural networks via self
  distillation. In: Proceedings of the IEEE/CVF International Conference on
  Computer Vision. pp. 3713--3722 (2019)

\bibitem{Zhang_2018_CVPR}
Zhang, X., Zhou, X., Lin, M., Sun, J.: {ShuffleNet}: An extremely efficient
  convolutional neural network for mobile devices. In: Proceedings of the IEEE
  Conference on Computer Vision and Pattern Recognition (2018)

\bibitem{zhu2020alleviating}
Zhu, W., Liao, H., Li, W., Li, W., Luo, J.: Alleviating the incompatibility
  between cross entropy loss and episode training for few-shot skin disease
  classification. In: International Conference on Medical Image Computing and
  Computer Assisted Intervention. pp. 330--339. Springer (2020)

\end{thebibliography}
%
% \begin{thebibliography}{8}
% \bibitem{ref_article1}
% Author, F.: Article title. Journal \textbf{2}(5), 99--110 (2016)

% \bibitem{ref_lncs1}
% Author, F., Author, S.: Title of a proceedings paper. In: Editor,
% F., Editor, S. (eds.) CONFERENCE 2016, LNCS, vol. 9999, pp. 1--13.
% Springer, Heidelberg (2016). \doi{10.10007/1234567890}

% \bibitem{ref_book1}
% Author, F., Author, S., Author, T.: Book title. 2nd edn. Publisher,
% Location (1999)

% \bibitem{ref_proc1}
% Author, A.-B.: Contribution title. In: 9th International Proceedings
% on Proceedings, pp. 1--2. Publisher, Location (2010)

% \bibitem{ref_url1}
% LNCS Homepage, \url{http://www.springer.com/lncs}. Last accessed 4
% Oct 2017
% \end{thebibliography}
%
%%%%%%%%%% Merge with supplemental materials %%%%%%%%%%
\newpage
%\widetext
\begin{center}
\textbf{\large Supplementary Material: Unsupervised Representation Learning Meets Pseudo-Label Supervised Self-Distillation: A New Approach to Rare Disease Classification}
\end{center}
%%%%%%%%%% Merge with supplemental materials %%%%%%%%%%
%%%%%%%%%% Prefix a "S" to all equations, figures, tables and reset the counter %%%%%%%%%%
\setcounter{equation}{0}
\setcounter{figure}{0}
\setcounter{table}{0}
\setcounter{page}{1}
\makeatletter
\renewcommand{\theequation}{S\arabic{equation}}
\renewcommand{\thefigure}{S\arabic{figure}}
\renewcommand{\thetable}{S\arabic{table}}
\renewcommand{\thepage}{S\arabic{page}}
%%%%%%%%%% Prefix a "S" to all equations, figures, tables and reset the counter %%%%%%%%%%

\begin{table}
\begin{center}
\caption{Performance comparison of alternative ways of applying the distilled student model (with ResNet-12 backbone and adaptive hard labels). ``Direct'' means directly using the student model $F'=f'_c(f_q')$, and ``LR'' means replacing $f'_c$ with a logistic regression classifier fit to the support set.
To validate that the difference in performance is not due to the exact type of the classifier, we also show the results of using a fully connected layer (FC) finetuned on the support set for the baseline model.
Data format: mean (standard deviation).}\label{tab:classifier}
\begin{adjustbox}{width=\columnwidth}
\begin{tabular}{l|l|c|c|c|c|c|c}
\hline
\multirow{2}{*}{Method} & \multirow{2}{*}{Classifier} & \multicolumn{2}{c}{($N$, $K$) = (3, 1)} & \multicolumn{2}{|c}{($N$, $K$) = (3, 3)} & \multicolumn{2}{|c}{($N$, $K$) = (3, 5)} \\
\cline{3-8}
& & Accuracy (\%)  & F1 score (\%)  & Accuracy (\%) & F1 score (\%)  & Accuracy (\%)  & F1 score (\%) \\
\hline
MoCo\_v1  \cite{he2020momentum} &LR & \textbf{61.90} (2.92) & \textbf{56.30} (1.48)  & \textbf{74.92} (2.96) & \textbf{69.50} (5.72)  & \textbf{79.01} (2.00) & \textbf{74.47} (3.03)   \\
                   (baseline)   &FC & 60.91 (2.19)& 55.71 (1.26)& 74.16 (3.77)& 68.59 (7.15)& 78.20 (3.37)& 74.39 (3.90)\\
\hline
{Hybrid-distill} & LR          & 62.86 (2.70)& 56.79 (4.38)& 74.45 (2.59)& 71.41 (4.14)& 80.55 (1.94)& 75.45 (1.62)\\
          (ours) & Direct          & \textbf{64.15} (2.86) & \textbf{61.01} (1.30) & \textbf{75.82} (2.47) & \textbf{73.34} (2.30) & \textbf{81.16} (2.60) & \textbf{77.35} (4.21)\\
                                    % & FC-finetune  & 64.71 (2.57)& 62.99 (1.05)& 76.43 (2.57)& 73.21 (1.05)& 81.49 (2.67)& 77.75 (4.07)\\
\hline
\end{tabular}
\end{adjustbox}
\end{center}
\end{table}

\begin{table}
    \begin{center}
    \caption{Accuracy comparison of four alternative designs of the pseudo labels (with ResNet-12 as backbone network).
    Data format: mean (standard deviation).}\label{tab:pseudo_label}
    \begin{adjustbox}{width=\columnwidth}
    \begin{tabular}{l|c|c|c|c|c|c}
        \hline
        \multirow{2}{*}{Pseudo label design}& \multicolumn{2}{c}{($N$, $K$) = (3, 1)} & \multicolumn{2}{|c}{($N$, $K$) = (3, 3)} & \multicolumn{2}{|c}{($N$, $K$) = (3, 5)} \\
        \cline{2-7}
        & Accuracy (\%)  & F1 score (\%)  & Accuracy (\%) & F1 score (\%)  & Accuracy (\%)  & F1 score (\%) \\
        \hline
        Soft  & 62.16 (4.23)& 56.25 (4.30)& 75.17 (3.19)& 71.48 (2.92)& 79.89 (2.46)& 74.61 (4.88) \\
        Hard  & 63.68 (3.00)& 58.69 (4.82)& 75.70 (2.90)& 71.75 (2.75)& 79.76 (1.27)& 75.68 (2.42) \\
        Adaptive soft & 62.26 (3.01)& 57.10 (4.12)& 75.45 (2.39)& 69.23 (4.26)& 79.74 (1.64)& 74.41 (3.51) \\
        Adaptive hard & \textbf{64.15} (2.86) & \textbf{61.01} (1.30) & \textbf{75.82} (2.47) & \textbf{73.34} (2.30) & \textbf{81.16} (2.60) & \textbf{77.35} (4.21)\\
        \hline
    \end{tabular}
    \end{adjustbox}
    \end{center}
\end{table}

\begin{table}
    \begin{center}
        \caption{Performance comparison of the baseline model using different backbones.
        Data format: mean (standard deviation).}\label{tab:backbone}
        \begin{adjustbox}{width=\columnwidth}
            \begin{tabular}{l|c|c|c|c|c|c}
            \hline
            \multirow{2}{*}{Backbone} & \multicolumn{2}{c}{($N$, $K$) = (3, 1)} & \multicolumn{2}{|c}{($N$, $K$) = (3, 3)} & \multicolumn{2}{|c}{($N$, $K$) = (3, 5)} \\
            \cline{2-7}
            & Accuracy (\%)  & F1 score (\%)  & Accuracy (\%) & F1 score (\%)  & Accuracy (\%)  & F1 score (\%) \\
            \hline
            4 conv blocks \cite{vinyals2016matching} & 53.36 (8.89) & 46.90 (6.34)  & 67.48 (6.06) & 59.77 (5.86)  & 69.94 (5.67) & 64.31 (3.15) \\
            MoblileNet\_v2 \cite{Sandler_2018_CVPR} & 52.03 (6.40)& 44.76 (4.90)& 57.21 (7.10)& 48.26 (5.83)& 61.92 (10.09)& 56.85 (7.33) \\
            ShuffleNet \cite{Zhang_2018_CVPR}  & 60.91 (2.19)& 55.71 (1.26)& 74.16 (3.77)& \textbf{69.59} (7.15)& 78.20 (3.37)& \textbf{75.39} (3.90) \\
            ResNet-12 \cite{tian2020rethinking} & \textbf{61.90} (2.92) & \textbf{56.30} (1.48)  & \textbf{74.92} (2.96) & 69.50 (5.72) & \textbf{79.01} (2.00) & 74.47 (3.03)  \\
            ResNet-18 \cite{tian2020rethinking} & 56.97 (1.15)& 48.27 (4.25)& 71.88 (4.83)& 65.35 (7.56)& 77.27 (4.01)& 71.80 (4.16) \\
            ResNet-50 \cite{tian2020rethinking} & 59.90 (1.60)& 48.08 (5.20)& 70.96 (1.74)& 63.10 (4.37)& 74.17 (1.99)& 68.66 (2.26) \\
            DenseNet-121 \cite{Huang_2017_CVPR} & 50.26 (4.38)& 44.03 (4.74)& 67.71 (0.71)& 59.22 (3.08)& 71.53 (1.79)& 66.54 (1.89) \\
            \hline
            \end{tabular}
        \end{adjustbox}
    \end{center}
\end{table}

\begin{table}[h]
\begin{center}
\caption{Evaluation results and comparison with SOTA FSL methods with the 4 conv blocks as backbone network.
Data format: mean (standard deviation).}\label{tab_isic_4conv}%of three random tasks $\mathcal{T}\sim p(\mathcal{T})$.
\begin{adjustbox}{width=\columnwidth}
\begin{threeparttable}
\begin{tabular}{l|c|c|c|c|c|c}
\hline
\multirow{2}{*}{Method} & \multicolumn{2}{c}{($N$, $K$) = (3, 1)} &  \multicolumn{2}{|c}{($N$, $K$) = (3, 3)} & \multicolumn{2}{|c}{($N$, $K$) = (3, 5)}  \\
\cline{2-7}
& Accuracy (\%)  & F1 score (\%)  & Accuracy (\%) & F1 score (\%)  & Accuracy (\%)  & F1 score (\%) \\
\hline
Training from scratch   & 35.91 (4.11) & 30.11 (1.73)  & 38.95 (4.03) & 33.24 (2.89)  & 43.13 (5.50) & 37.57 (1.32)  \\
\hdashline
     $\triangleright$ SML
\hfill MAML \cite{finn2017model}   & 46.20 (4.22) & 39.84 (2.53)  & 51.56 (3.03) & 45.38 (2.28)  & 55.73 (3.62) & 49.51 (2.68) \\
\hfill RelationNet \cite{sung2018learning}   & 41.11 (4.72) & 36.47 (4.60)  & 42.63 (4.57) & 38.49 (4.77)  & 49.98 (5.81) & 42.86 (4.81) \\
\hfill ProtoNets \cite{snell2017prototypical} & 34.97 (1.69) & 28.31 (2.05)  & 36.38 (1.91) & 31.65 (2.73)  & 37.95 (2.98) & 34.08 (2.40)  \\
\hfill \hfill DAML \cite{li2020difficulty}   & 46.70 (2.74) & 40.96 (5.11)  & 53.91 (3.23) & 44.80 (6.22)  & 56.11 (2.39) & 53.29 (2.42)  \\
\hdashline
     $\triangleright$ UML
\hfill UMTRA \cite{khodadadeh2019unsupervised}  & 41.45 (2.47) & 35.88 (0.26)  & 45.76 (2.78) & 39.07 (2.35)  & 54.16 (2.20) & 49.41 (2.55) \\
\hfill CACTUs-MAML \cite{hsu2018unsupervised}  & 39.45 (3.42) & 33.09 (2.60)  & 42.20 (2.47) & 38.70 (3.80)  & 45.31 (1.69) & 39.96 (3.46) \\
\hfill CACTUs-ProtoNets \cite{hsu2018unsupervised}   & 33.24 (2.12) & 29.69 (0.84)  & 35.08 (1.70) & 30.95 (2.69)  & 36.86 (1.95) & 32.76 (3.10)  \\
\hdashline
     $\triangleright$ SRL
\hfill SRL-simple \cite{tian2020rethinking}   & 45.08 (6.59) & 37.15 (4.91)  & 57.50 (4.48) & 45.56 (0.62)  & 60.62 (4.34) & 53.98 (3.29) \\
\hfill SRL-distil \cite{tian2020rethinking}    & 50.14 (3.74) & 43.76 (2.54)  & 58.20 (4.15) & 52.24 (3.06)  & 62.07 (5.11) & 58.71 (5.64) \\
\hdashline
     $\triangleright$ URL
\hfill SimCLR \cite{chen2020simple}         & 48.07 (6.46) & 44.91 (7.94)  & 60.78 (4.18) & 55.32 (3.52)  & 66.25 (5.36) & 61.75 (4.28) \\
\hfill MoCo\_v2 \cite{chen2020improved}    & 48.88 (8.31) & 44.45 (7.16)  & 59.71 (2.54) & 52.68 (3.75)  & 62.91 (2.04) & 57.43 (2.95) \\
\hfill MoCo\_v1 (baseline) \cite{he2020momentum}      & 53.36 (8.89) & 46.90 (6.34)  & 67.48 (6.06) & 59.77 (5.86)  & 69.94 (5.67) & 64.31 (3.15) \\
\hfill \textbf{Hybrid distil (ours)}  & \textbf{54.72} (8.46)& \textbf{47.28} (4.45)  & \textbf{69.14} (5.36) & \textbf{62.56} (5.84)  & \textbf{70.26} (4.97) & \textbf{65.89} (4.60) \\
\hline

\end{tabular}
% \begin{tablenotes}\footnotesize
%     \item[a] Despite our efforts, we are unable to tune ProtoNets to yield comparable results with other methods.
% \end{tablenotes}
\end{threeparttable}
\end{adjustbox}
\end{center}
\end{table}

\end{document}